\documentclass[letterpaper, 10 pt, conference]{ieeeconf}  

\IEEEoverridecommandlockouts                              

\usepackage{graphics} 
\usepackage{cite}
\usepackage{amsmath,amsfonts}
\usepackage{algorithmic}
\usepackage{algorithm}
\usepackage{array}
\usepackage{textcomp}
\usepackage{stfloats}
\usepackage{url}
\usepackage{verbatim}
\usepackage{graphicx}
\usepackage{cite}
\usepackage{multirow}
\usepackage{threeparttable}
\usepackage{csquotes}

\title{\LARGE \bf
UniMSF: A Unified Multi-Sensor Fusion Framework for Intelligent Transportation System Global Localization*
}

\author{Wei Liu$^{1}$, Jiaqi Zhu$^{1}$, Guirong Zhuo$^{1}$, Wufei Fu$^{1}$, Zonglin Meng$^{2}$, Yishi Lu$^{3}$, \\Min Hua$^{4}$, Feng Qiao$^{5}$, You Li$^{6}$, Yi He$^{7}$ and Lu Xiong$^{1}$
\thanks{*This work was supported by the National Natural Science Foundation of China (No. 52325212), the National Key Research and Development Program of China (No. 2022YFE0117100), and Shanghai Tongyu Automobile Technology Intelligent Vehicle By-Wire Chassis Joint Laboratory.}
\thanks{$^{1}$School of Automotive Studies, Tongji University, Shanghai 201804, China. (Corresponding author: Jiaqi Zhu: e-mail:{~\tt\small zjq970220@tongji.edu.cn}).}%
\thanks{$^{2}$Department of Civil and Environmental Engineering, University of California, Los Angeles, Los Angeles, CA, U.S.A.}%
\thanks{$^{3}$Shanghai Gongji Technology Co., Ltd, Shanghai 201804, China.}%
\thanks{$^{4}$Department of Mechanical
Engineering, the University of Birmingham, B15 2TT, UK.}%
\thanks{$^{5}$Institute for Automotive Engineering (ika), RWTH Aachen University, Aachen, 52062, Germany.}%
\thanks{$^{6}$State Key Laboratory of Surveying Mapping and Remote Sensing Wuhan University, Wuhan, China.}%
\thanks{$^{7}$Intelligent Transportation System Research Center, Wuhan University of Technology, Wuhan, China.}%
}

\begin{document}

\maketitle
\thispagestyle{empty}
\pagestyle{empty}

\begin{abstract}

Intelligent transportation systems (ITS) localization is of significant importance as it provides fundamental position and orientation for autonomous operations like intelligent vehicles. Integrating diverse and complementary sensors such as global navigation satellite system (GNSS) and 4D-radar can provide scalable and reliable global localization. Nevertheless, multi-sensor fusion encounters challenges including heterogeneity and time-varying uncertainty in measurements. Consequently, developing a reliable and unified multi-sensor framework remains challenging. In this paper, we introduce UniMSF, a comprehensive multi-sensor fusion localization framework for ITS, utilizing factor graphs. By integrating a multi-sensor fusion front-end, alongside outlier detection\&noise model estimation, and a factor graph optimization back-end, this framework accomplishes efficient fusion and ensures accurate localization for ITS. Specifically, in the multi-sensor fusion front-end module, we tackle the measurement heterogeneity among different modality sensors and establish effective measurement models. Reliable outlier detection and data-driven online noise estimation methods ensure that back-end optimization is immune to interference from outlier measurements. In addition, integrating multi-sensor observations via factor graph optimization offers the advantage of \enquote{plug and play}. Notably, our framework features high modularity and is seamlessly adapted to various sensor configurations. We demonstrate the effectiveness of the proposed framework through real vehicle tests by tightly integrating GNSS pseudorange and carrier phase information with IMU, and 4D-radar.

\end{abstract}

\section{INTRODUCTION}

Localization technology is integral to our daily lives, not only featured in devices like smartwatches and smartphones but also essential for the safe operation of intelligent transportation systems (ITS) including autonomous ground vehicles, aircraft, and watercraft~\cite{shen2024novel}. For ITS, scalability and reliability are the basic yet key concerns of existing localization solutions. For instance, in challenging scenarios such as tunnels or urban canyons, the reliability of global navigation satellite system (GNSS), one of the most fundamental sensors for outdoor autonomous vehicle systems, substantially diminishes. This reduction in reliability is primarily due to multipath effects and non-line-of-sight (NLOS) reception. To ensure reliable and scalable localization in complex environments, it is imperative to leverage available multi-modality sensors in ITS. Specifically, widely used heterogeneous sensors~\cite{xia2022autonomous}, including the Inertial Measurement Unit (IMU), camera, LiDAR, and millimeter-wave radar, provide significant opportunities for integration, thereby enhancing the reliability of GNSS-based localization. 

The IMU is widely adopted with GNSS due to its cost-effectiveness and environmental robustness. Two principal methods of fusion, namely loose coupling and tight coupling, are predominantly employed. Loose coupling combines position and velocity data from real-time kinematic (RTK) GNSS, while tight coupling uses raw GNSS measurements (pseudorange, Doppler shift, and carrier phase) for better reliability in challenging environments~\cite{shen2024novel}. Recently, factor graph-based tight coupling of raw GNSS data and IMU has shown better localization performance than filter-based methods~\cite{shen2024novel}. Single-point positioning (SPP), relying on pseudorange data, achieves only meter-level accuracy, insufficient for autonomous driving. RTK-GNSS, using carrier phase data, attains centimeter-level accuracy but requires auxiliary base stations and challenging real-time integer ambiguity resolution.
The time-difference carrier phase (TDCP) circumvents the necessity for integer ambiguity resolution and facilitates accurate relative positioning~\cite{bai2023performance, suzuki2022gnss, bai2022time}. Its superior performance has prompted the increased adoption of this technique for processing carrier phase data. The principal challenge associated with TDCP arises from cycle slips in the carrier phase under complex environmental conditions, leading to notable declines in localization accuracy~\cite{bai2023performance}. Additionally, noise and bias in the IMU would contribute to error drift, particularly during extended periods of compromised or absent GNSS signals.


Cameras and LiDAR excel in precise local localization, especially in complex, feature-rich environments, and complement GNSS well.
The excellent localization performance achieved through the tight integration of visual or LiDAR data with pseudorange~\cite{cao2022gvins}, RTK-GNSS~\cite{wang2023give, li2023enhancing}, or TDCP~\cite{beuchert2023factor} underscores their superior complementary attributes. However, camera and LiDAR systems may experience performance degradation under adverse weather or challenging lighting conditions. In these scenarios, the recently developed 4D-radar has demonstrated robust performance~\cite{abu2023radar}, with its full potential yet to be fully realized. Additionally, the presence of sparse and noisy point clouds poses further challenges.

Although the integration of sensors such as the IMU, camera, and LiDAR with GNSS offers numerous benefits, it simultaneously poses a variety of challenges. Firstly, to optimize the synergistic potential of their distinctive attributes and ensure robust global localization accuracy, the development of a unified and scalable framework facilitating the seamless integration of multiple sensors with GNSS is urgent for ITS. In addition, considerable heterogeneity arises in the measurements owing to the distinct operational modalities inherent to each sensor. To maximize the utilization of each sensor's capabilities, consideration should be given to the selection of data for fusion. Moreover, in intricate and dynamic scenarios, the variability or degradation of individual sensor errors can complicate the accurate determination of their covariances. Therefore, dynamic adjustment of fusion strategy is essential, contingent upon the quality of information, to alleviate the impact of noise or potential outlier measurements on the fusion outcomes.

To address these challenges, in this paper, we propose UniMSF, a novel multi-sensor unified fusion framework for ITS localization based on factor graph optimization. It serves as a general framework to integrate GNSS with other modality sensors. Besides fully utilizing the complementary properties of heterogeneous sensors, it dynamically adjusts weights based on information quality for efficient fusion. UniMSF comprises a multi-sensor fusion front-end, outlier detection\&noise model estimation, and a factor graph optimization back-end. Specifically, in the multi-sensor fusion front-end module, we consider the characteristics of different sensors to establish effective measurements. Secondly, reliable outlier detection methods are designed to mitigate the influence of outlier measurements. Adaptive fusion is also achieved through the developed noise estimation module. Finally, leveraging factor graph optimization, the integration of multi-sensor observations offers the advantages of ``plug and play" and scalability. The contributions of this paper are as follows:
\begin{itemize}

\item To the best of our knowledge, this is the first paper to facilitate the scalable integration of GNSS with diverse modality sensors, thus improving the reliability of global localization in ITS. The framework, based on factor graph optimization, provides the advantage of scalability.
\item In the proposed framework, two distinct categories of GNSS factors are delineated: pseudorange for global measurements and TDCP for enhanced reliability in relative measurements. Our study extensively explores the integration of these factors with other sensor modalities. Additionally, the inclusion of outlier detection\&noise model estimation modules ensures the robustness and adaptability of the fusion process.
\item Utilizing the autonomous driving system as an experimental platform, we demonstrate the effectiveness of the proposed framework, exemplified by its seamless integration of pseudorange, TDCP, IMU, and 4D-radar data sources under complex driving scenarios.

\end{itemize}

The remainder of this paper is organized as follows: Section~\ref{sec:system overview} outlines the architecture of the proposed UniMSF. Section~\ref{sec:Case Study} details a case study entitled Radar-UniMSF, an innovative fusion strategy for the 4D-radar enhanced IMU/Pseudorange/TDCP tight coupling system, aimed at intelligent vehicle localization. Experimental validation and discussions are presented in Section~\ref{sec:Experiment}. The conclusion is provided in Section~\ref{sec:CONCLUSIONS}.

\section{System Overview}\label{sec:system overview}

\begin{figure*}[hbt]
    \centering
    \includegraphics[width=1\linewidth]{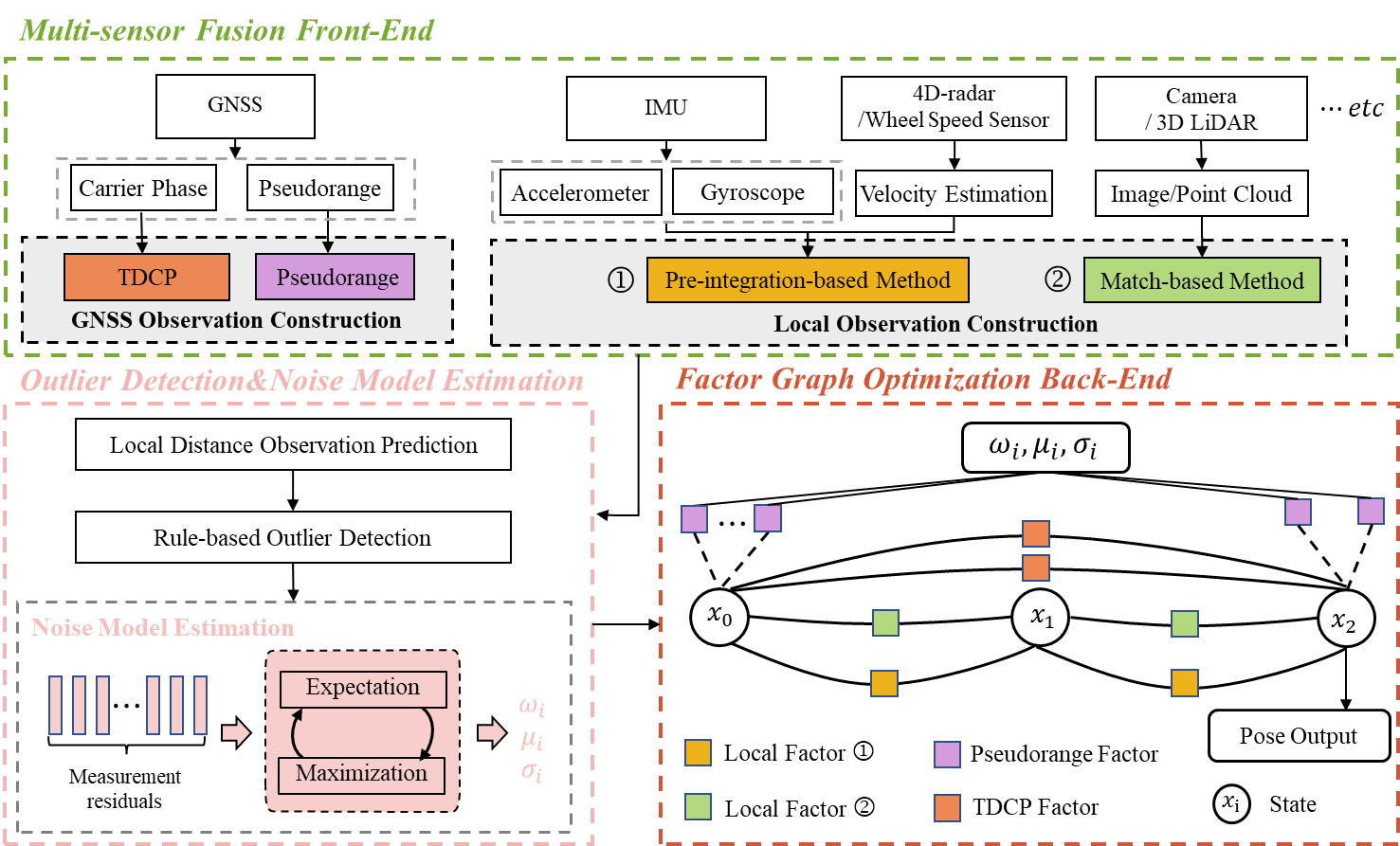}
    \vspace{-20pt}
    \caption{Overall structure of the proposed UniMSF, comprising three main modules: multi-sensor fusion front-end, outlier detection\&noise model estimation, and factor graph optimization back-end.}
    \label{fig:system_overview}
\end{figure*}

This paper innovatively introduces UniMSF, a novel framework for multi-sensor fusion based on factor graph optimization, aiming to enhance the localization performance for ITS. The overall architecture of UniMSF is shown in Fig.~\ref{fig:system_overview}. The system incorporates inputs from various sensors, including the camera, LiDAR, 4D-radar, IMU (accelerometer and gyroscope), and GNSS receivers (pseudorange and carrier phase). These sensors serve as examples, and the framework is adaptable to incorporate additional sensor modalities as needed. The framework comprises three primary modules: a multi-sensor fusion front-end, outlier detection\&noise model estimation, and a factor graph optimization back-end.

Within the multi-sensor fusion front-end module, we meticulously account for the heterogeneous characteristics exhibited by diverse sensors to establish robust measurements. Specifically, GNSS contributes two distinct types of measurements: pseudorange, providing global observations, and TDCP, delivering more dependable relative measurements. 
For measurements from sensors of other modalities, we offer illustrative examples of common sensor types to provide clarity. For sensors such as the IMU and wheel speed sensors that provide vehicle acceleration, angular velocity, and velocity measurements, we adopt the pre-integration-based method~\cite{qin2018vins} to generate local measurements. For sensors like the cameras or LiDAR, which supply image or point cloud measurements, local measurements are derived using match-based methods as described in previous works~\cite{cao2022gvins,li2023enhancing,beuchert2023factor}. Although emerging 4D-radar also offers point cloud measurements, the sparse and noisy nature of these point clouds diminishes the effectiveness of match-based methods. To address this issue, we propose to estimate ego-velocity based on 4D-radar data and utilize the pre-integration-based method to generate local measurements.

The outlier detection\&noise model estimation module assesses the reliability of both local and global measurements generated by the front-end module. Onboard sensors like the IMU remain largely unaffected by environmental factors and demonstrate high reliability. Conversely, external sensors such as the camera, LiDAR, GNSS, and others are considerably influenced by environmental conditions. 
Formulating a dependable outlier processing strategy is crucial for ensuring accurate and reliable localization performance. Thus, we elegantly devise a rule-based outlier detection method to identify and discard measurements characterized by significant errors. Subsequently, we estimate the confidence of the remaining measurements by modeling the noise using a data-driven approach with a Gaussian Mixture Model (GMM). This module serves to estimate the parameters of the GMM, allowing for dynamic adjustment of fusion weights based on information quality, thus ensuring efficient fusion.

The factor graph optimization back-end module is responsible for generating factors using filtered measurements and their corresponding weights. Multiple factors can be incorporated into the factor graph to facilitate graph optimization. Notably, the factor graph framework is characterized by its ``plug and play" nature and inherent scalability. By integrating suitable front-end components and factors, the framework enables flexible integration of additional sensors such as magnetometers, event cameras, and more.

\section{Case Study: Radar-UniMSF}\label{sec:Case Study}
Building upon UniMSF, this section presents a case study of Radar-UniMSF, an efficient fusion algorithm for autonomous driving localization utilizing a 4D-radar-enhanced IMU/Pseudorange/TDCP tight coupling system. 4D-radar, an emerging sensor, presents notable cost efficiencies and maintains stability in adverse environmental conditions, including rain, snow, fog, and low light. Both academia and industry have demonstrated considerable interest in this technology. However, there have been limited studies conducted on its application in vehicle localization. To bridge this gap,  this section focuses on the implications of tightly integrating 4D-radar with raw GNSS data for vehicle localization. It provides an extensive description of the framework proposed in this paper. To the best of our knowledge, this is the first work in the field of vehicle localization systems to achieve tight integration of IMU, 4D-radar, GNSS pseudorange, and carrier phase data.

In this paper, we denote the ENU frame as $\left ( \cdot \right ) ^{n} $, the ECEF frame as $\left ( \cdot \right ) ^{e} $, and the 4D-radar frame as $\left ( \cdot \right ) ^{r} $. The body frame, which coincides with the IMU frame, is represented by $\left ( \cdot \right ) ^{b} $, while $\left ( \cdot \right ) ^{b_{k}} $ signifies the $k$-th body frame. The relative pose transformation from $\left ( \cdot \right ) ^{b}$ to $\left ( \cdot \right ) ^{w} $ is denoted by $\mathbf{T}_{b}^{w} \in SE\left ( 3 \right ) $. Additionally, we utilize the rotation matrix $\mathbf{R}$ and the Hamilton quaternion $\mathbf{q}$ to represent rotations.

\subsection{Multi-sensor Fusion Front-End}\label{sec:front-end}

Radar-UniMSF receives inputs from the 4D-radar, IMU and GNSS receivers. In the multi-sensor fusion front-end, we preintegrate the IMU data. For the 4D-radar, we propose a RANSAC-based method to estimate the ego velocity and perform the velocity pre-integration. Observation models for GNSS pseudorange and carrier phase measurements are constructed, incorporating elevation filtering and error compensation. Notably, TDCP is utilized for carrier phase measurements.

\subsubsection{IMU Pre-integration}\label{sec:IMU-Pre}
To pre-integrate IMU measurements, we adopt the methods outlined in VINS-Mono~\cite{qin2018vins}. Based on the measured acceleration and angular velocity within the time interval $\left [ t_{k},t_{k+1} \right ] $, we compute the pre-integration terms as follows:
\begin{align}
    \boldsymbol{\alpha} _{b_{k+1}}^{b_{k}} &=\iint_{t\in \left [ t_{k}, t_{k+1}\right ] } \left ( \mathbf{R}_{b_{t}}^{b_{k}}\left ( \mathbf{a}_{t}-\mathbf{b}_{a_{t}}  \right )   \right ) \mathrm{d}t ^{2} \notag \\
    \boldsymbol{\beta}  _{b_{k+1}}^{b_{k}}&=\int_{t\in \left [ t_{k}, t_{k+1}\right ] } \left ( \mathbf{R}_{b_{t}}^{b_{k}}\left ( \mathbf{a}_{t}-\mathbf{b}_{a_{t}}  \right )   \right ) \mathrm{d}t \notag \\
    \boldsymbol{\gamma} _{b_{k+1}}^{b_{k}}&=\int_{t\in \left [ t_{k}, t_{k+1}\right ] } \frac{1}{2}\boldsymbol{\Omega}  \left ( \boldsymbol{\omega} _{t}-\mathbf{b}_{\omega _{t}}  \right )\boldsymbol{\gamma} _{b_{t}}^{b_{k}}  \mathrm{d}t
    \label{eqn:1}
\end{align} 
where $\boldsymbol{\alpha} _{b_{k+1}}^{b_{k}}$, $\boldsymbol{\beta}  _{b_{k+1}}^{b_{k}}$, and $\boldsymbol{\gamma} _{b_{k+1}}^{b_{k}}$  are the position, velocity, and attitude pre-integration terms, respectively. $\mathbf{a}_{t}$ and $\boldsymbol{\omega} _{t}$ are the IMU measurements at the time $t$. $\mathbf{b}_{a_{t}}$ and $\mathbf{b}_{\omega _{t}}$ represent the accelerometer bias and the gyroscope bias, respectively. 

\subsubsection{4D-radar Velocity Pre-integration}\label{sec:Vel-Pre}
Unlike traditional 3D LiDAR systems, the 4D-radar not only provides 3D environmental measurements but also offers Doppler velocity for every point in the scene. This unique feature provides crucial data for accurate vehicle localization. Here, we propose a method for estimating the vehicle's ego velocity using 4D-radar measurements. Subsequently, we develop a position pre-integration model based on this velocity measurement. The Doppler velocity measurement ${v} _{r}$ represents the relative radial velocity between the measurement point and the radar, and it is calculated as follows:
\begin{align}
    {v} _{r} = \frac{\mathbf{p} ^{r} }{\left \| \mathbf{p} ^{r} \right \| }\cdot\left ( \mathbf{v} _{ego}^{r}-\mathbf{v} _{obj}   \right ) 
    =\overrightarrow{\mathbf{d} _{p} }  \cdot  \left ( \mathbf{v} _{ego}^{r}-\mathbf{v} _{obj}  \right ) 
    \label{eqn:2}
\end{align} 
where $\mathbf{p} ^{r} =\left [ \begin{matrix} x_{p}^{r}
  &y_{p}^{r}  &z_{p}^{r}
\end{matrix}  \right ] $ is the position of the measurement point. $\overrightarrow{\mathbf{d}_{p} }  =\left [ \begin{matrix} \frac{x_{p}^{r}}{l} 
  &\frac{y_{p}^{r}}{l}   &\frac{z_{p}^{r}}{l} 
\end{matrix}  \right ] = \left [ \begin{matrix} \cos {\theta}  _{x} 
  &\cos {\theta}  _{y}   &\cos {\theta}  _{z} 
\end{matrix} \right ]$. $l = \sqrt{{x_{p}^{r}}^2 + {y_{p}^{r}}^2 + {z_{p}^{r}}^2} $. $\mathbf{v} _{obj} $ represents the velocity of the measurement point. $\mathbf{v} _{ego}^{r} =\left [ \begin{matrix}v_{ego_{x}}^{r} 
  &v_{ego_{y}}^{r}   &v_{ego_{z}}^{r} 
\end{matrix} \right ]^{T}  $ represents the velocity of the vehicle in 4D-radar frame. From Equation~(\ref{eqn:2}), it is evident that both $\mathbf{v} _{ego}^{r}$ and $\mathbf{v} _{obj}$ are unknown. Given that static target points have $\mathbf{v} _{obj} = 0$. Therefore, it is more advantageous to utilize solely static points for ego-velocity estimation. The RANSAC~\cite{fischler1981random} algorithm is employed to extract static points from a frame of the 4D-radar point cloud. Subsequently, a set of Equation (\ref{eqn:3}) is constructed based on the extracted static points, and the least squares algorithm is then used to estimate the 2D ego velocity $\mathbf{v} _{ego_{2D}}^{r}=\left [ \begin{matrix}v_{ego_{x}}^{r}
  &v_{ego_{y}}^{r}
\end{matrix} \right ]^{T} $.
\begin{align}
    \left [ \begin{matrix}\cos \theta _{x} 
        &\cos \theta _{y} 
        \end{matrix} \right ] \mathbf{v} _{ego_{2D}}^{r}={v} _{r}
        \label{eqn:3}
\end{align}

To mitigate the impact of estimation noise, we solely consider the longitudinal velocity and disregard lateral and vertical velocities, setting them to zero. The final vehicle velocity measurement is ${\hat{\mathbf{v}} } _{t}=\left [ \begin{matrix}v_{ego_{x}}^{b}
  &0 & 0
\end{matrix} \right ]^{T} $. $\mathbf{v} _{ego}^{b} = \mathbf{R} _{r}^{b} \mathbf{v} _{ego}^{r}$. $\mathbf{R} _{r}^{b}$ denotes the rotation matrix from the 4D-radar frame to the body frame. In the time interval $\left [ t_{k},t_{k+1} \right ] $, we can formulate the position pre-integration term as follows:
\begin{align}
    \boldsymbol{\eta}  _{b_{k+1}}^{b_{k}}=\int_{t\in \left [ t_{k}, t_{k+1}\right ] } \left ( \mathbf{R}_{b_{t}}^{b_{k}}  \hat{\mathbf{v}} _{t}\right )\mathrm{d}t 
    \label{eqn:4}
\end{align} 
where $\mathbf{R}_{b_{t}}^{b_{k}}$ equals to $\mathbf{R}\left ( \boldsymbol{\gamma} _{b_{t}}^{b_{k}} \right )$, which is obtained from IMU rotation pre-integration term $\boldsymbol{\gamma} _{b_{t}}^{b_{k}}$.

\subsubsection{Pseudorange Measurement}\label{sec:Pseudorange Measurement}

The GNSS receiver supplies raw measurements such as ephemeris, pseudorange, and carrier phase data. Notably, pseudorange information offers a global constraint on the position, aiding in error mitigation and accumulation prevention. At time $t_{k}$, the pseudorange measurement for satellite $s_{j}$ is modeled as follows~\cite{kaplan2017understanding}:
\begin{align}
    \rho _{k,j} &=\left \| \mathbf{p}_{s_{j},k }^{e}- \mathbf{p}_{r_{k}}^{e} \right \| +\delta t_{k} -\delta t_{s_{j}, k}+\delta \rho _{kn,k}+\delta \rho _{kp,k} \notag \\
    &+\delta _{s_{j},k}^{sag}
    \label{eqn:5}
\end{align}
where $\mathbf{p}_{s_{j},k }^{e} = \begin{bmatrix}x_{s_{j},k }^{e}
  &y_{s_{j},k }^{e} &z_{s_{j},k }^{e}
\end{bmatrix}$ is the position of satellite $s_{j}$ in ECEF frame. $ \mathbf{p}_{r_{k}}^{e} = \begin{bmatrix}x_{r_{k}}^{e}
  &y_{r_{k}}^{e} &z_{r_{k} }^{e}
\end{bmatrix}$ stands for the position of the GNSS receiver antenna in ECEF frame. $\delta t_{k}$ and $\delta t_{s_{j},k}$ denote the receiver clock error and the satellite clock error, respectively, the unit is $m$. Among them, $\mathbf{p}_{s_{j},k }^{e}$ and $\delta t_{s_{j},k}$ are calculated by ephemeris information. $\mathbf{p}_{r_{k}}^{e}$ and $\delta t_{k}$ are estimated as unknowns. $\delta \rho _{kn,k}$ and $\delta \rho _{kp,k}$ are atmospheric delays and ionospheric delays, respectively. $\delta _{s_{j},k}^{sag}$ denotes the Sagnac effect of the Earth's rotation. These three values are modeled and compensated using the method in~\cite{takasu2009development}.

\subsubsection{TDCP Measurement}\label{sec:TDCP Measurement}
In contrast to pseudorange, carrier phase data provides significantly higher accuracy. Achieving centimeter-level  is feasible once the integer ambiguity in the carrier phase is resolved correctly. However, this task is challenging, particularly in scenarios characterized by tall buildings, occlusions, and other obstructions. At time $t_{k}$, the carrier phase measurement for satellite $s_{j}$ is expressed as described in~\cite{kaplan2017understanding}:
\begin{align}
    \varphi _{k,j} &=\left \| \mathbf{p}_{s_{j},k }^{e}- \mathbf{p}_{r_{k}}^{e} \right \| +\delta t_{k} -\delta t_{s_{j}, k}+\delta \rho _{kn,k}-\delta \rho _{kp,k} \notag \\
    &+\lambda N_{s_{j},k} 
    \label{eqn:6}
\end{align}
where $N_{s_{j},k}$ and $\lambda$ denote the integer ambiguity and wavelength, respectively. From Equation (\ref{eqn:6}), it is evident that the integer ambiguity of the carrier phase can be eliminated via time difference, provided there is no cycle slip. This approach circumvents the challenging task of integer ambiguity resolution. Hence, we employ the time difference to formulate the carrier phase measurements. Taking the adjacent epochs $t_{k-1}$ and $t_{k}$ as an example, the TDCP measurements of satellite $s_{j}$ are as follows:
\begin{align}
 \bigtriangleup \varphi _{k,k-1}^{j} & = \varphi _{k,j} - \varphi _{k-1,j} \notag \\
&=\left \| \mathbf{p}_{s_{j},k }^{e}- \mathbf{p}_{r_{k}}^{e} \right \| -\left \| \mathbf{p}_{s_{j},k-1 }^{e}- \mathbf{p}_{r_{k-1}}^{e} \right \| \notag \\
&+\delta t_{k} -\delta t_{k-1} 
\label{eqn:7}
\end{align}

Equation~(\ref{eqn:7}) illustrates that TDCP measurements offer a robust constraint on the receiver position at different time points, assuming no cycle slips occur. Consequently, the development of a dependable cycle slip detection module is paramount.

\subsection{Outlier Detection\&Noise Model Estimation}\label{sec:Outlier and noise}
\subsubsection{Outlier Detection based on the Integration of Doppler Frequency Shift}\label{sec:Outlier detection}
To ensure reliable cycle slip detection in carrier phase, we adopt the method based on Doppler frequency shift integration as outlined in~\cite{bai2023performance}. This is because Doppler frequency shift serves as an independent measurement unaffected by cycle slips~\cite{zhao2020high}. For the carrier phase observations of satellite $s_{j}$ at epochs $t_{k-1}$ and $t_{k}$, the constructed detection metric is as follows:
\begin{align}
\varepsilon &=\left | \bigtriangleup \varphi _{k,k-1}^{j} -\lambda \int_{t_{k-1}}^{t_{k}}D_{t,j}dt   \right | \notag \\
&=\left | \bigtriangleup \varphi _{k,k-1}^{j} -\lambda\frac{D_{k-1,j}+D_{k,j}}{2}\bigtriangleup t   \right |  
\label{eqn:8}
\end{align}

\begin{figure}[t]
    \centering
    \vspace{8pt}
    \includegraphics[width=1\linewidth]{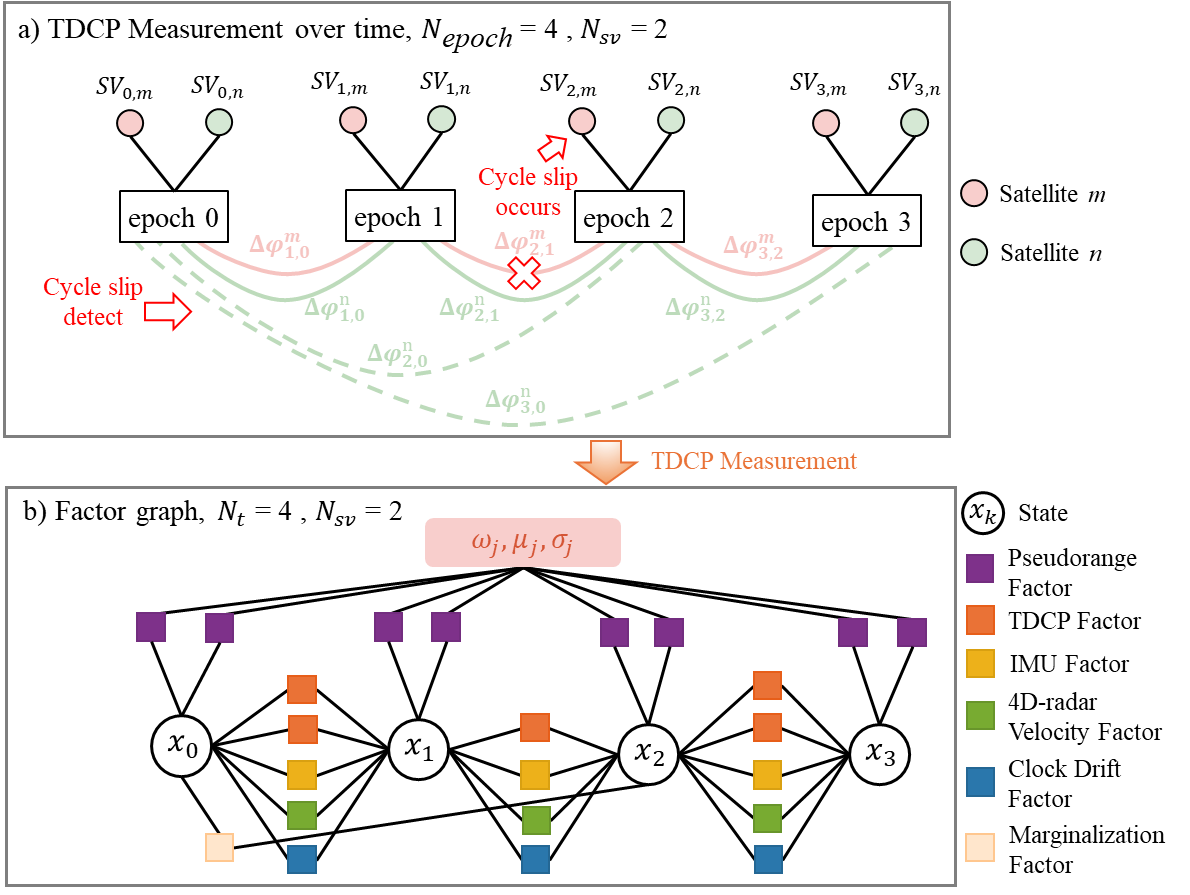}
    \vspace{-20pt}
    \caption{Illustration of factor graph for the proposed Radar-UniMSF. a) The construction process of TDCP measurements, using an example of 2 satellites observed over 4 continuous epochs. The solid lines represent the TDCPs between adjacent epochs that are utilized, while dashed lines indicate those across multiple epochs, which are optional. b) The factor graph framework of Radar-UniMSF.}
    \label{fig:factor_graph}
\end{figure}

If $\varepsilon$ is less than the defined threshold, we infer the absence of cycle slips. Subsequently, TDCP measurements, as depicted in Fig.~\ref{fig:factor_graph}a, is utilized. TDCP is constructed for epochs where no cycle slips are identified. Although minor cycle slips may persist undetected, their incorporation with additional multi-sensor data can enhance the smoothness of the estimation results and alleviate their impact.

\subsubsection{Noise Model Estimation}\label{sec:noise estimation}

GNSS data, particularly pseudorange measurements, are profoundly influenced by environmental factors. As environmental conditions fluctuate, so does the noise model associated with pseudorange measurements. Consequently, obtaining precise covariance estimates for pseudorange observations presents a challenge. In this paper, we employ the EM algorithm~\cite{pfeifer2019expectation} to estimate the noise parameters of pseudorange measurements. The noise distribution is modeled using a GMM, comprising multiple weighted Gaussian components. It is expressed as follows:
\begin{align}
\mathbf{P}\left ( \boldsymbol{z}_{i}\mid \boldsymbol{x}_{i}  \right )\propto  {\textstyle \sum_{j}\omega _{j}\cdot \lambda\left ( \mu _{j},\sigma _{j}^{2}  \right )  }
    \label{eqn:9}
\end{align}
where, $\boldsymbol{z}_{i}$ and $\boldsymbol{x}_{i}$ represent the \(i\)-th measurement and state, respectively. $\omega_{j}$ denotes the weight of the \(j\)-th Gaussian component, while $\mu_{j}$ and $\sigma_{j}^{2}$ stand for the mean and variance of this Gaussian component, respectively. To estimate these noise parameters online, we employ the EM algorithm, which uses a sequence of measurement residuals as input. Firstly, the pseudorange residuals $e_{k,j}$ can be obtained as shown in Equation (\ref{eqn:10}) by substituting the initial estimate of the current state into Equation (\ref{eqn:5}). By iteratively performing the E-step and M-step, we estimate the noise parameters. In our implementation, we opt for the exact sum-mixture approach and fix the number of Gaussian components at 2.
\begin{align}
 e_{k,j} &=\left \| \mathbf{p}_{s_{j},k }^{e}- \mathbf{p}_{r_{k}}^{e} \right \| +\delta t_{k} -\delta t_{s_{j}, k}+\delta \rho _{kn,k} \notag  \\
    &+\delta \rho _{kp,k}+\delta _{s_{j},k}^{sag}-Z_{\rho _{k,j}} 
    \label{eqn:10}
\end{align}

\subsection{Factor Graph Optimization Back-end}\label{sec:back end}
The factor graph, a probabilistic graphical model consisting of nodes and factors, serves to estimate target variables by maximizing the posterior probability. We employ the factor graph to jointly optimize constraints from the IMU, 4D-radar, and GNSS. As illustrated in Fig.~\ref{fig:factor_graph}b, the factor graph of the proposed Radar-UniMSF consists of IMU factor, 4D-radar velocity factor, clock drift factor, pseudorange factor, TDCP factor, and marginalization factor. The state vector $\boldsymbol{\chi}$ is summarized as follows:
\begin{align}
\boldsymbol{\chi}  &=\left [ \boldsymbol{x}_{0} ,\boldsymbol{x}_{1},\cdots ,\boldsymbol{x}_{k},\cdots,\boldsymbol{x}_{L}   \right ]  \notag \\
  \boldsymbol{x}_{k} &= \left [ \mathbf{p}_{b_{k} }^{n},\mathbf{v}_{b_{k} }^{n},\mathbf{q}_{b_{k} }^{n},\mathbf{b}_{a }^{b_{k}},\mathbf{b}_{\omega  }^{b_{k}},\delta t_{k},\dot{\delta }t_{k} \right ],k\in \left [ 0,L \right ]
    \label{eqn:11}
\end{align} 
where $\boldsymbol{\chi}$ is the states in the sliding window, and the window size is $L$. $\boldsymbol{x}_{k}$ stands for the state in time $t_{k}$, which includes: the position $\mathbf{p}_{b_{k}}^{n}$, velocity $\mathbf{v}_{b_{k}}^{n}$, orientation $\mathbf{q}_{b_{k}}^{n}$, accelerometer bias $\mathbf{b}_{a}^{b_{k}}$, gyroscope bias $\mathbf{b}_{\omega }^{b_{k}}$, receiver clock bias $\delta t_{k}$, and receiver clock drifting rate $\dot{\delta} t_{k}$.

The error function of the proposed Radar-UniMSF is formulated as follows:
\begin{align}
\boldsymbol{\chi}^{\ast } &=arg\min_{\boldsymbol{x}} \left \{\left \| \mathbf{r}_{P}\left ( \boldsymbol{\chi} \right )   \right \|^{2}  \notag  \right.\\
&\left. +{\textstyle\sum_{k=0}^{L-1}}\left \| \mathbf{r}_{I}\left (\widetilde{\mathbf{Z}}_{b_{k+1}}^{b_{k}} , \boldsymbol{\chi} \right )  \right \|_{\boldsymbol{\sum} _{I} }^{2} \notag  \right.\\ 
&\left.  +{\textstyle\sum_{k=0}^{L-1}}\left \| \mathbf{r}_{V}\left (\widetilde{\mathbf{Z}}_{b_{k+1}}^{b_{k}} , \boldsymbol{\chi} \right ) \right \|_{\boldsymbol{\sum} _{V} }^{2}  \notag  \right.\\
&\left.  +{\textstyle\sum_{k=0}^{L-1}}\left \| \mathbf{r}_{C}\left (\widetilde{\mathbf{Z}}_{b_{k+1}}^{b_{k}} , \boldsymbol{\chi} \right ) \right \|_{\boldsymbol{\sum} _{C} }^{2}  \notag  \right.\\
&\left.  +{\textstyle\sum_{k=0}^{L-1}}\left \| \mathbf{r}_{T}\left (\widetilde{\mathbf{Z}}_{b_{k+1}}^{b_{k}} , \boldsymbol{\chi} \right ) \right \|_{\boldsymbol{\sum} _{T} }^{2}  \notag  \right.\\
&\left.  +{\textstyle\sum_{k=0}^{L}}\left \| \mathbf{r}_{pr}\left (\widetilde{\mathbf{Z}}_{r_{k},s_{j}} , \boldsymbol{\chi} \right ) \right \| _{GMM}^{2} 
\right \}
   \label{eqn:12}
\end{align}
where $\mathbf{r}_{P}\left ( \boldsymbol{\chi} \right )$ is the prior information obtained from marginalization. $\mathbf{r}_{I}\left (\widetilde{\mathbf{Z}}_{b_{k+1}}^{b_{k}} , \boldsymbol{\chi} \right )$, $\mathbf{r}_{V}\left (\widetilde{\mathbf{Z}}_{b_{k+1}}^{b_{k}} , \boldsymbol{\chi} \right )$, $\mathbf{r}_{C}\left (\widetilde{\mathbf{Z}}_{b_{k+1}}^{b_{k}} , \boldsymbol{\chi} \right )$, $\mathbf{r}_{T}\left (\widetilde{\mathbf{Z}}_{b_{k+1}}^{b_{k}} , \boldsymbol{\chi} \right )$, and $\mathbf{r}_{pr}\left (\widetilde{\mathbf{Z}}_{r_{k},s_{j}} , \boldsymbol{\chi} \right )$
represent the IMU, 4D-radar velocity, clock drift, TDCP, and pseudorange residuals, respectively. $\boldsymbol{\sum} _{I}$, $\boldsymbol{\sum} _{V}$, $\boldsymbol{\sum} _{C}$, and $\boldsymbol{\sum} _{T}$ represent the covariance matrices of the IMU and velocity pre-integration factors, clock drift factor, and TDCP factor, which are set based on the Gaussian model. The noise of the pseudorange factor is modeled by GMM, whose parameters are set via the estimation results in Section~\ref{sec:noise estimation}. To address the non-linear optimization problem, we employ the Levenberg-Marquardt algorithm in Ceres Solver~\cite{agarwal2012ceres}. The specific models of above factors are given as follows.

\subsubsection{IMU Factor}\label{sec:imu factor}
We will not delve into the intricacies of constructing the IMU pre-integration factors, as the methodology mirrors that outlined in~\cite{qin2018vins}.

\subsubsection{4D-radar Velocity Factor}\label{sec:velocity factor}
In the time interval $[t_{k},t_{k+1}]$, the position pre-integration term is constructed using the 4D-radar-based ego-velocity estimation, which is calculated in Section~\ref{sec:Vel-Pre}. The residual is as follows:
\begin{align}
\mathbf{r}_{V}\left (\widetilde{\mathbf{Z}}_{b_{k+1}}^{b_{k}} , \boldsymbol{\chi} \right )
= \mathbf{R}_{n}^{b_{k}} \left (\mathbf{p}_{b_{k+1}}^{n}-\mathbf{p}_{b_{k}}^{n}\right )-\boldsymbol{\eta} _{b_{k+1}}^{b_{k}}
   \label{eqn:13}
\end{align}

\subsubsection{TDCP Factor}\label{sec:TDCP factor}
To mitigate the adverse effects of cycle slips, only TDCP measurements detected through cycle slip detection are incorporated into the factor graph. In theory, longer connection of TDCP measurements, such as dashed TDCP measurements depicted in Fig.~\ref{fig:factor_graph}a, offer enhanced error reduction. However, this typically results in increased computational overhead. To accommodate real-time constraints, only neighboring TDCP measurements are considered. During the time interval $[t_{k},t_{k+1}]$, the TDCP residual for satellite $s_{j}$, concerning the state and TDCP measurement computed in Section~\ref{sec:TDCP Measurement}, is expressed as:
\begin{align}
\mathbf{r}_{T}\left (\widetilde{\mathbf{Z}}_{b_{k+1}}^{b_{k}} , \boldsymbol{\chi} \right )
&=\left \| \mathbf{p}_{s_{j},k+1 }^{e}- \mathbf{p}_{r_{k+1}}^{e} \right \| -\left \| \mathbf{p}_{s_{j},k }^{e}- \mathbf{p}_{r_{k}}^{e} \right \| \notag \\
&+\delta t_{k+1} -\delta t_{k} - \bigtriangleup \varphi _{k+1,k}^{j} 
   \label{eqn:14}
\end{align}

\subsubsection{Pseudorange Factor}\label{sec:pseudorange factor}
We add all pseudorange measurements that meet the elevation threshold requirement to the factor graph. In this paper, we estimate the position of the vehicle in the ENU frame: $ \mathbf{p}_{b_{k}}^{n} = \begin{bmatrix}x_{b_{k} }^{n}
  &y_{b_{k}}^{n} &z_{b_{k}}^{n}
\end{bmatrix}$. Therefore, a transform of the ENU frame with the ECEF frame is required, and the arm $l_{g}^{b}$ between the GNSS antenna and the IMU also should be considered. In time $t_{k}$, The pseudorange residual for satellite $s_{j}$ that relates the states and pseudorange measurements is formulated as follows:
\begin{align}
    \mathbf{r}_{pr}\left (\widetilde{\mathbf{Z}}_{r_{k},s_{j}} , \boldsymbol{\chi} \right )  &=\left \| \mathbf{p}_{s_{j},k }^{e}- \mathbf{p}_{r_{k}}^{e} \right \| +\delta t_{k} -\delta t_{s_{j}, k}+\delta \rho _{kn,k} \notag  \\
    &+\delta \rho _{kp,k}+\delta _{s_{j},k}^{sag}-Z_{\rho _{k,j}} \notag  \\
    \mathbf{p}_{r_{k}}^{e} &=\mathbf{T}_{n}^{e}\left ( \ \mathbf{p}_{b_{k}}^{n} + \mathbf{R}_{b_{k}}^{n}\boldsymbol{l}_{g}^{b} \right )
    \label{eqn:15}
\end{align}

In addition, the unknown clock error of the receiver is also an estimated variable, which often drifts at a certain rate. Hence, we employ  the constant clock error drift (CCED) model~\cite{kaplan2017understanding}, depicted below:
\begin{align}
    \mathbf{r}_{C}\left (\widetilde{\mathbf{Z}}_{b_{k+1}}^{b_{k}} , \boldsymbol{\chi} \right )  =\begin{bmatrix}\delta t_{k} +\dot{\delta t}_{k}\triangle t 
 \\ \dot{\delta t}_{k}
\end{bmatrix}-\begin{bmatrix}\delta t_{k+1}
 \\\dot{\delta t}_{k+1}
\end{bmatrix}
    \label{eqn:16}
\end{align}

\section{EXPERIMENTAL RESULTS AND DISCUSSION}\label{sec:Experiment}
This section provides an overview of the experimental platform, followed by a discussion of experimental results.

\begin{figure}[b]
    \centering
    \includegraphics[width=0.95\linewidth]{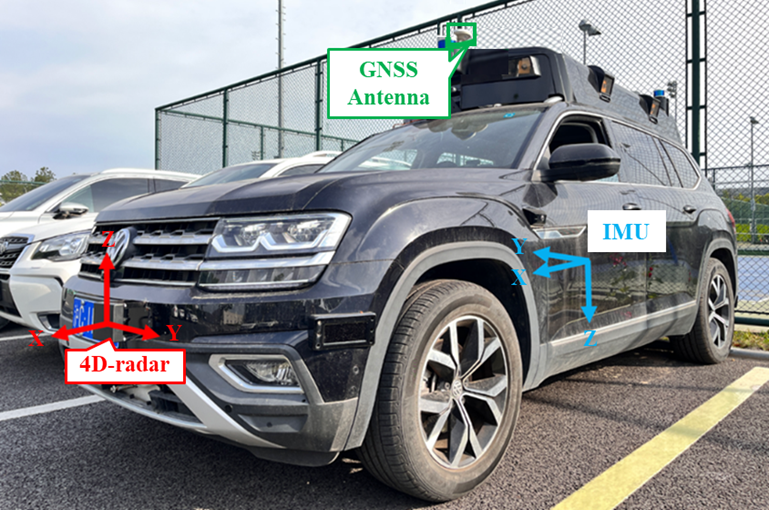}
    \vspace{-10pt}
    \caption{Experimental platform for data collection.}
    \label{fig:platform}
\end{figure}

\subsection{Experiment Environment and Setup}
\begin{figure}[]
    \centering
    \vspace{5pt}
    \includegraphics[width=0.9\linewidth]{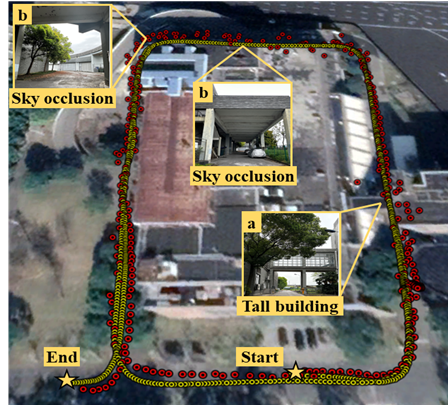}
    \vspace{-10pt}
    \caption{Trajectory and experimental scenarios. (a) Tall building and brief sky occlusion region; (b) Long duration sky occlusion region.}
\label{fig:scenarios}
\end{figure}

Due to the absence of existing open-source datasets containing 4D-radar, GNSS pseudorange, and carrier phase information concurrently, we employ the vehicle platform depicted in Fig.~\ref{fig:platform} for data collection at Tongji University's Jiading campus to validate the proposed algorithm. The experimental vehicle is outfitted with an Oculii 4D-radar operating at 15 Hz. The MEMS-IMU ASM330LHH provides three-axis acceleration and angular velocity measurements at 100 Hz. The automotive-grade U-blox F9P GNSS receiver provides satellite observation data (BeiDou) at 1 Hz. Additionally, the OxTS-RT3000 system provides high-precision vehicle states as ground truth (GT). We have driven the car for 2 laps in this scene. The yellow line depicts the ground truth trajectory of the vehicle, while yellow stars mark the starting and ending points of the route. The red line shows the trajectory of SPP using only GNSS pseudorange information, implemented with RTKLIB~\cite{takasu2009development}. The SPP results highlight the challenges by GNSS in this scenario. In scenes (a) and (b) of Fig.~\ref{fig:scenarios}, the accuracy of the GNSS is affected by tall buildings and sky obstacles, resulting in poor performance.

To evaluate the effectiveness of the proposed algorithm, we compare with the following algorithms:
\begin{itemize}

\item IPT: A  method based on a factor graph with tight integration of the IMU, pseudorange, and carrier phase~\cite{bai2023performance}. To take into account the time cost, we only add the TDCP constraints for adjacent epochs.
\item Radar-VMSF (Vanilla version of the proposed method): In addition to IMU, pseudorange, and carrier phase data, the inclusion of velocity information estimated by 4D-radar further enhances localization performance. In this scenario, 4D-radar velocity data is utilized to construct the observation through pre-integration.
\item Radar-UniMSF (Proposed method): This is the comprehensive version of our algorithm. It incorporates online estimation of the pseudorange noise model into Radar-VMSF, which mitigates the effects of GNSS outliers. Unlike IPT and Radar-VMSF, where the pseudorange noise is modeled as a Gaussian distribution with covariance calculated based on parameters such as signal-to-noise ratio and satellite elevation angle~\cite{takasu2009development}, our method offers enhanced accuracy and robustness.
\end{itemize}

In the local ENU frame, we evaluate the accuracy of the three methods mentioned above. The performance metrics used for the quantitative analysis are the Mean Absolute Error (MAE) and the Root Mean Square Error (RMSE). In addition, we present the results of the attitude estimation.

\subsection{Experimental Evaluation}
\begin{figure}[]
    \vspace{-10pt}
    \centering
    \includegraphics[width=0.95\linewidth]{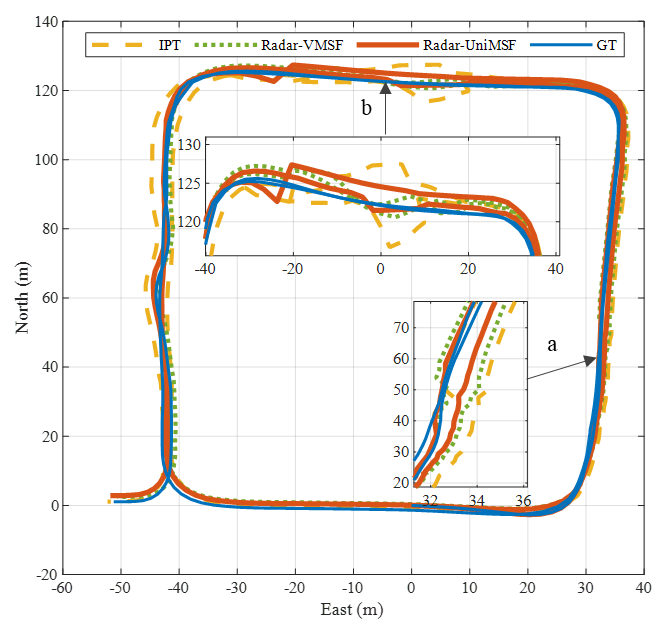}
    \vspace{-10pt}
    \caption{Trajectories of the three methods: IPT (yellow), Radar-VMSF (green) and Radar-UniMSF (red). Blue line indicates GT. The local trajectory in scenes (a) and (b) are also zoomed in.}
    \label{fig:trajectory}
\end{figure}

\begin{figure}[]
    \vspace{5pt}
    \centering
    \includegraphics[width=0.98\linewidth]{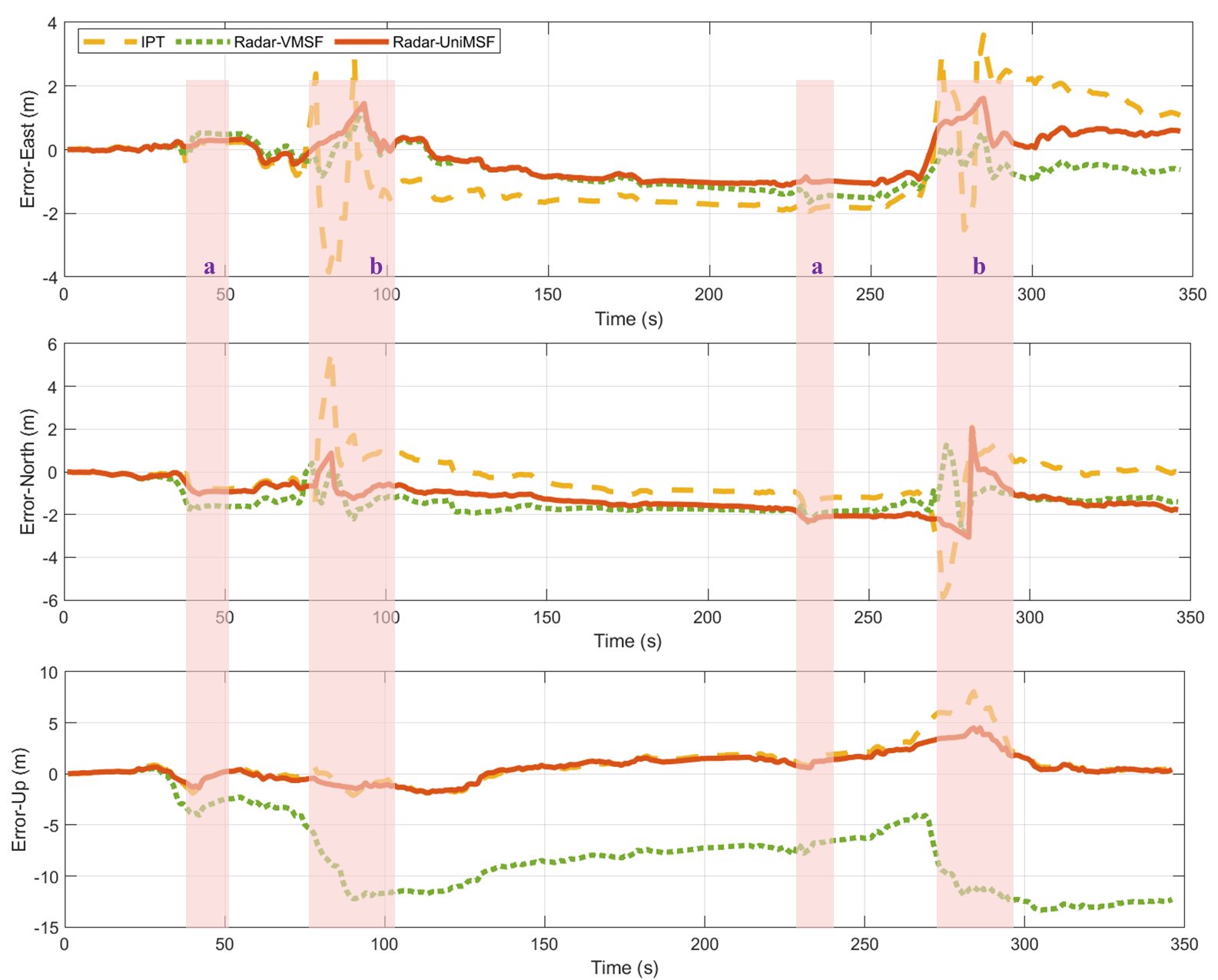}
    \vspace{-10pt}
    \caption{Position errors of the tested methods.}
    \label{fig:pos_err}
\end{figure}

\begin{figure}[]
    \vspace{-10pt}
    \centering
    \includegraphics[width=0.98\linewidth]{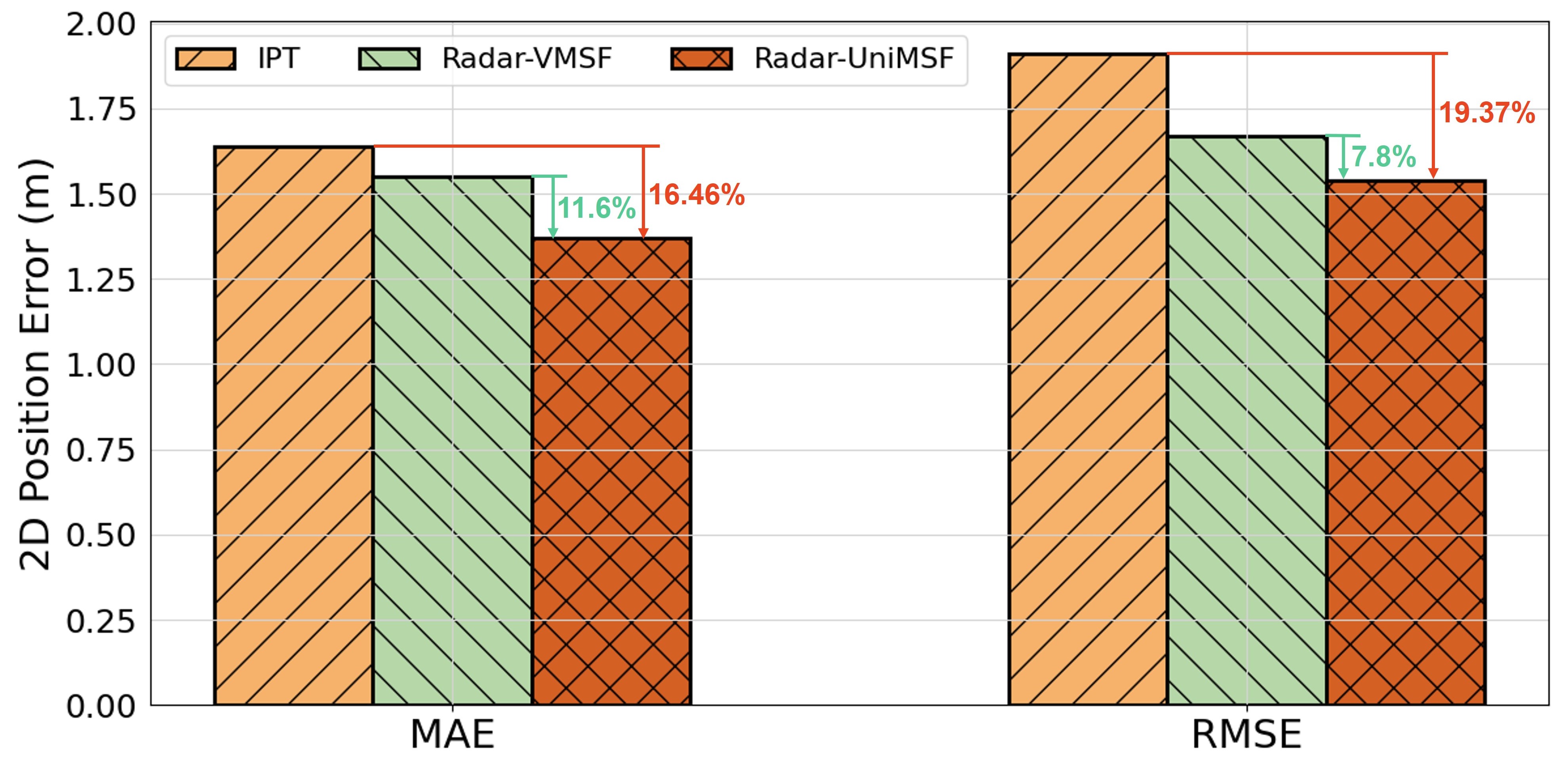}
    \vspace{-10pt}
    \caption{2D Position errors of the evaluated methods.}
    \label{fig:com}
\end{figure}

\begin{figure}[t]
    \vspace{5pt}
    \centering
    \includegraphics[width=0.96\linewidth]{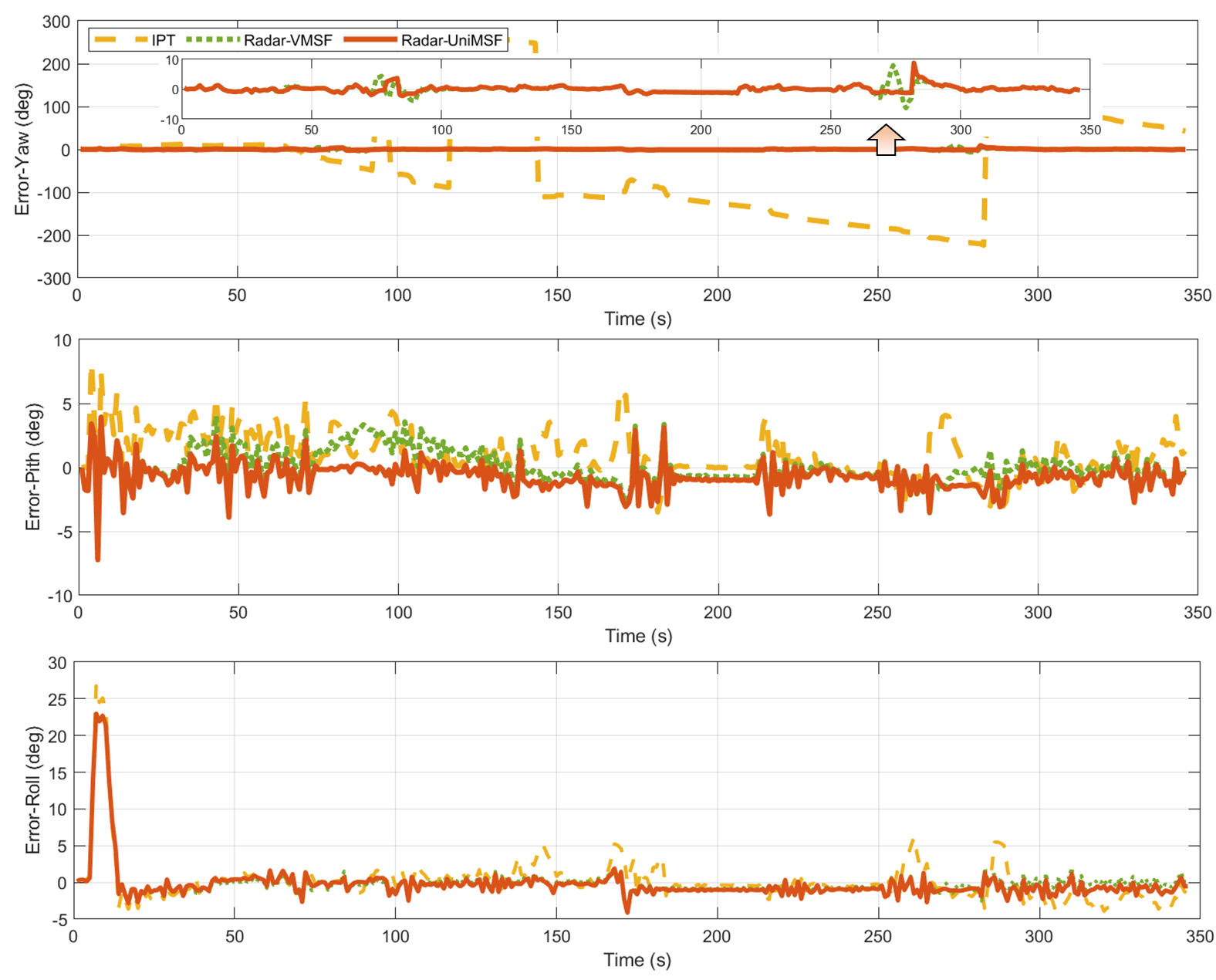}
    \vspace{-10pt}
    \caption{Attitude errors of the tested methods. For better visualisation, the yaw errors of Radar-VMSF and Radar-UniMSF are displayed enlarged.}
    \label{fig:attitude_err}
\end{figure}

\begin{figure}[]
    \centering
    \includegraphics[width=0.96\linewidth]{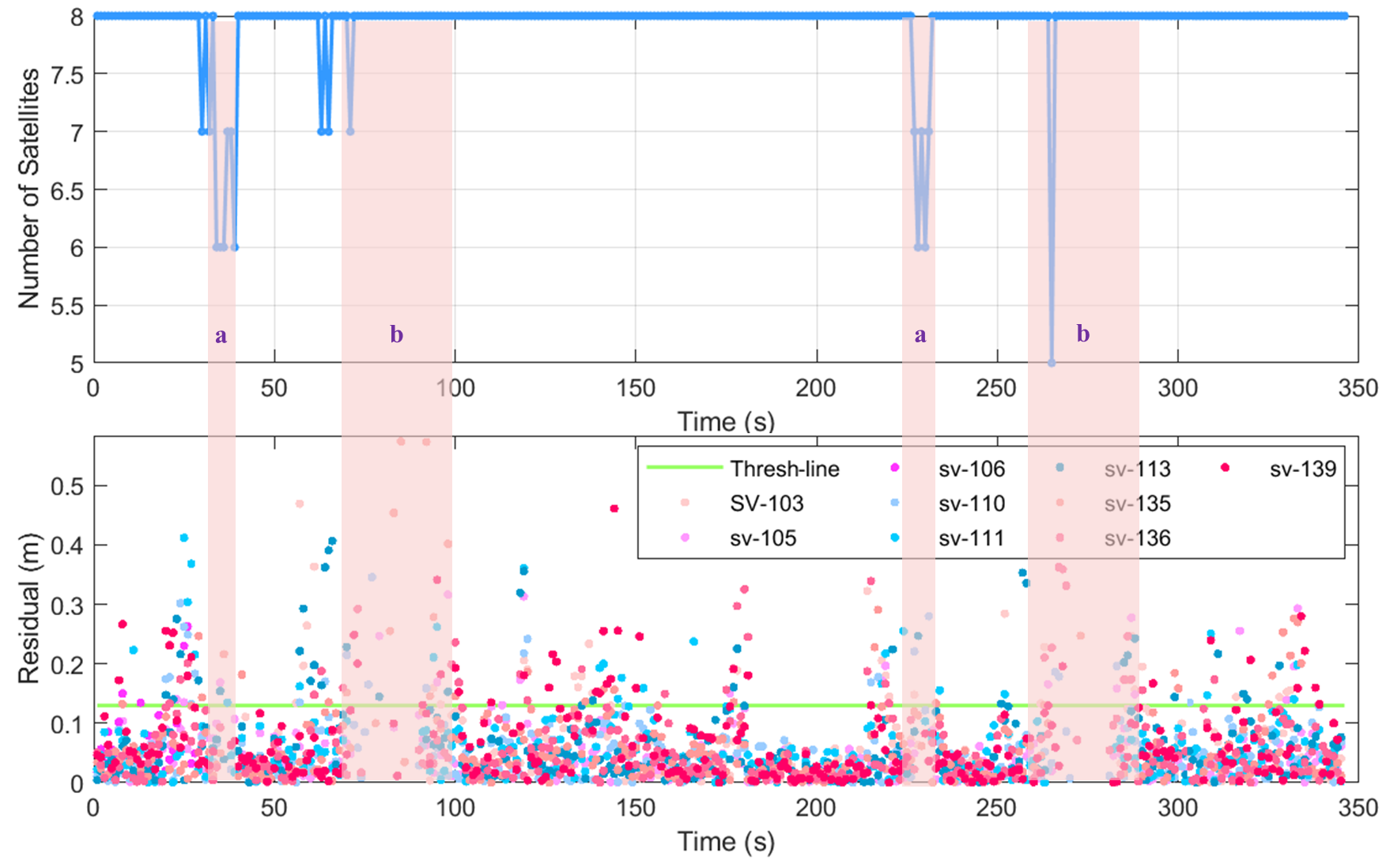}
    \vspace{-10pt}
    \caption{Number of satellites (top) and residuals between TDCP and Doppler frequency shift integration (bottom). In this case, TDCP with residuals greater than the threshold are rejected.}
    \label{fig:sv}
\end{figure}

Fig.~\ref{fig:trajectory} illustrates the trajectories of the three methods. Firstly, we observe a significant improvement in accuracy with IPT compared to SPP in Fig.~\ref{fig:scenarios}. This is evidence that the addition of TDCP can effectively improve the accuracy of GNSS. Furthermore, both the proposed Radar-VMSF and Radar-UniMSF achieve better performance than IPT, proving the effectiveness of integrating other sensor support into GNSS/IMU system. Among these, Radar-UniMSF is closest to GT. Fig.~\ref{fig:pos_err} shows the absolute localization errors in the East, North, and Up directions, respectively. Radar-UniMSF exhibits the smallest positioning errors in all three axes, achieving the best performance. The results of the quantitative analysis of positioning accuracy are summarized in Fig.~\ref{fig:com}. Overall, compared to the baseline IPT algorithm, the Radar-UniMSF proposed in this paper reduces MAE by 0.27 $m$ and improves accuracy by 19.37\% in RMSE.
Notably, the attitude error results in Fig.~\ref{fig:attitude_err} show a significant improvement based on Radar-UniMSF, particularly for yaw angle. This also has major advantages in improving positioning accuracy. These results confirm the effectiveness of the tightly coupled multi-sensor proposed in this paper.


In addition, we perform a specific analysis according to the scenario. Fig.~\ref{fig:sv} shows that the minimum number of satellites observed is 5. This meets the requirement of at least 4 satellites for GNSS. However, in scenarios (a) and (b), sky occlusion leads to outlier GNSS pseudorange measurements. On the other hand, there are fewer qualified TDCP measurements below the threshold, as shown at the bottom of Fig.~\ref{fig:sv}. This results in poor IPT performance. Moreover, outlier pseudorange observations lead to large trajectory jumps. As shown in Fig.~\ref{fig:trajectory}, Radar-VMSF mitigates these jumps with the help of 4D-radar velocity information. Nevertheless, the trajectory remains non-smooth. Radar-UniMSF introduces online estimation of the pseudorange noise, resulting in a smoother trajectory. Compared to Radar-VMSF, the accuracy is improved by 7.8\% in RMSE, which demonstrates the advantages of online noise estimation.

Interestingly, although the proposed Radar-UniMSF outperforms the other algorithms in several indicators, its performance is slightly worse than Radar-VMSF in passing through scene (b) in the second round. From the bottom of Fig.~\ref{fig:sv}, it is evident that there are currently few TDCP observations available. Therefore, localization is mainly based on the integrated IMU and 4D-radar velocity. The East and North positioning errors slowly drift due to the existence of small yaw angle errors until available TDCP observations are obtained. In the future, we will introduce sensors that can provide attitude constraints to enhance the accuracy of yaw angle, such as cameras.

\section{CONCLUSIONS}\label{sec:CONCLUSIONS}

This paper proposes UniMSF, a unified multi-sensor fusion framework for ITS localization based on factor graph optimization. It is a unified and scalable framework that integrates available modality sensors. The front-end module establishes appropriate measurements for heterogeneous sensors to fully exploit positioning potential. The outlier detection and noise estimation module dynamically adjusts weights based on information quality to achieve efficient fusion. Finally, factor graph optimization is introduced to integrate multi-sensor observations. We also provide an example of tightly integrating the pseudorange and carrier phase information with IMU, 4D-radar. The effectiveness is validated through real vehicle experiments in challenging GNSS scenarios. This paper provides insights into the design of multi-sensor fusion localization for ITS. Future research warrants the integration of deep learning-based methods to estimate the weight of each sensor measurement.

\addtolength{\textheight}{-12cm}   


\bibliographystyle{IEEEtran}

\begin{thebibliography}{10}

\bibitem{shen2024novel}
Z.~Shen, X.~Li, X.~Wang, Z.~Wu, X.~Li, Y.~Zhou, and S.~Li, ``A novel factor
  graph framework for tightly coupled gnss/ins integration with carrier-phase
  ambiguity resolution,'' \emph{IEEE Transactions on Intelligent
  Transportation Systems}, 2024.


\bibitem{xia2022autonomous}
X.~Xia, E.~Hashemi, L.~Xiong, and A.~Khajepour, ``Autonomous vehicle kinematics
  and dynamics synthesis for sideslip angle estimation based on consensus
  kalman filter,'' \emph{IEEE Transactions on Control Systems Technology},
  vol.~31, no.~1, pp. 179--192, 2022.


\bibitem{bai2023performance}
S.~Bai, J.~Lai, P.~Lyu, Y.~Cen, X.~Sun, and B.~Wang, ``Performance enhancement
  of tightly coupled gnss/imu integration based on factor graph with robust
  tdcp loop closure,'' \emph{IEEE Transactions on Intelligent Transportation
  Systems}, 2023.


\bibitem{suzuki2022gnss}
T.~Suzuki, ``Gnss odometry: Precise trajectory estimation based on carrier
  phase cycle slip estimation,'' \emph{IEEE Robotics and Automation Letters},
  vol.~7, no.~3, pp. 7319--7326, 2022.


\bibitem{bai2022time}
X.~Bai, W.~Wen, and L.-T. Hsu, ``Time-correlated window-carrier-phase-aided
  gnss positioning using factor graph optimization for urban positioning,''
  \emph{IEEE Transactions on Aerospace and Electronic Systems}, vol.~58,
  no.~4, pp. 3370--3384, 2022.


\bibitem{cao2022gvins}
S.~Cao, X.~Lu, and S.~Shen, ``Gvins: Tightly coupled gnss--visual--inertial
  fusion for smooth and consistent state estimation,'' \emph{IEEE Transactions
  on Robotics}, vol.~38, no.~4, pp. 2004--2021, 2022.


\bibitem{wang2023give}
X.~Wang, X.~Li, H.~Chang, S.~Li, Z.~Shen, and Y.~Zhou, ``Give: A tightly
  coupled rtk-inertial-visual state estimator for robust and precise
  positioning,'' \emph{IEEE Transactions on Instrumentation and Measurement},
  2023.


\bibitem{li2023enhancing}
X.~Li, S.~Wang, S.~Li, Y.~Zhou, C.~Xia, and Z.~Shen, ``Enhancing rtk
  performance in urban environments by tightly integrating ins and lidar,''
  \emph{IEEE Transactions on Vehicular Technology}, vol.~72, no.~8, pp.
  9845--9856, 2023.


\bibitem{beuchert2023factor}
J.~Beuchert, M.~Camurri, and M.~Fallon, ``Factor graph fusion of raw gnss
  sensing with imu and lidar for precise robot localization without a base
  station,'' in \emph{2023 IEEE International Conference on Robotics and
  Automation (ICRA)}, 2023, pp. 8415--8421.


\bibitem{abu2023radar}
N.~J. Abu-Alrub and N.~A. Rawashdeh, ``Radar odometry for autonomous ground
  vehicles: A survey of methods and datasets,'' \emph{IEEE Transactions on
  Intelligent Vehicles}, 2023.


\bibitem{qin2018vins}
T.~Qin, P.~Li, and S.~Shen, ``Vins-mono: A robust and versatile monocular
  visual-inertial state estimator,'' \emph{IEEE Transactions on Robotics},
  vol.~34, no.~4, pp. 1004--1020, 2018.


\bibitem{fischler1981random}
M.~A. Fischler and R.~C. Bolles, ``Random sample consensus: a paradigm for
  model fitting with applications to image analysis and automated
  cartography,'' \emph{Communications of the ACM}, vol.~24, no.~6, pp.
  381--395, 1981.


\bibitem{kaplan2017understanding}
E.~D. Kaplan and C.~Hegarty, \emph{Understanding gps/gnss: principles and
  applications}.\hskip 1em plus 0.5em minus 0.4em\relax Artech house, 2017.


\bibitem{takasu2009development}
T.~Takasu and A.~Yasuda, ``Development of the low-cost rtk-gps receiver with an
  open source program package rtklib,'' in \emph{International symposium on
  GPS/GNSS}, vol.~1, 2009, pp. 1--6.


\bibitem{zhao2020high}
J.~Zhao, M.~Hern{\'a}ndez-Pajares, Z.~Li, L.~Wang, and H.~Yuan, ``High-rate
  doppler-aided cycle slip detection and repair method for low-cost
  single-frequency receivers,'' \emph{Gps solutions}, vol.~24, pp. 1--13, 2020.


\bibitem{pfeifer2019expectation}
T.~Pfeifer and P.~Protzel, ``Expectation-maximization for adaptive mixture
  models in graph optimization,'' in \emph{2019 international conference on
  robotics and automation (ICRA)}, 2019, pp. 3151--3157.


\bibitem{agarwal2012ceres}
S.~Agarwal and K.~Mierle, ``Ceres solver,'' 2012.


\end{thebibliography}

\end{document}